\documentclass{article}
\usepackage{ijcai23}

\usepackage{times}
\usepackage{soul}
\usepackage{url}
\usepackage[hidelinks]{hyperref}
\usepackage[utf8]{inputenc}
\usepackage[small]{caption}
\usepackage{graphicx}
\usepackage{amsmath}
\usepackage{amsthm}
\usepackage{booktabs}
\usepackage{algorithm}
\usepackage{algorithmic}
\usepackage[switch]{lineno}
\usepackage{multirow}
\usepackage{pifont}
\usepackage{utfsym}

\usepackage{amssymb}

\title{A Dual Semantic-Aware Recurrent Global-Adaptive Network for Vision-and-Language Navigation}

\author{
Liuyi Wang$^{1}$\and
Zongtao He$^1$\and
Jiagui Tang$^1$\and
Ronghao Dang$^1$\and
Naijia Wang$^1$\and
Chengju Liu$^{1,2}$\footnote{corresponding author}\And
Qijun Chen$^{1*}$
\affiliations
$^1$Tongji University, Shanghai, China\\
$^2$Tongji Artificial Intelligence (Suzhou) Research Institute, Suzhou, China\\
\emails
\{wly, xingchen327, 2130701, dangronghao, 2030715, liuchengju, qjchen\}@tongji.edu.cn
}
\begin{document}

\maketitle

\begin{abstract}
Vision-and-Language Navigation (VLN) is a realistic but challenging task that requires an agent to locate the target region using verbal and visual cues. While significant advancements have been achieved recently, there are still two broad limitations: (1) The explicit information mining for significant guiding semantics concealed in both vision and language is still under-explored; (2) The previously structured map method provides the average historical appearance of visited nodes, while it ignores distinctive contributions of various images and potent information retention in the reasoning process. This work proposes a dual semantic-aware recurrent global-adaptive network (DSRG) to address the above problems. First, DSRG proposes an instruction-guidance linguistic module (IGL) and an appearance-semantics visual module (ASV) for boosting vision and language semantic learning respectively. For the memory mechanism, a global adaptive aggregation module (GAA) is devised for explicit panoramic observation fusion, and a recurrent memory fusion module (RMF) is introduced to supply implicit temporal hidden states. Extensive experimental results on the R2R and REVERIE datasets demonstrate that our method achieves better performance than existing methods. Code is available at \url{https://github.com/CrystalSixone/DSRG}. 
\end{abstract}

\section{Introduction}
\label{sec_introduction}
As an important application in human-robot interaction, the vision-language navigation (VLN) task~\cite{anderson2018vision} has attracted considerable attention. It is a crucial but challenging task that requires an embodied agent to reach the required locations in unstructured environments only based on visual observations and given verbal instructions. The main issue relies on how to effectively mine and exploit the high-level context connotation hidden in the plentiful features of vision and language so that they can serve as better guiding elements in the serial navigation process.
\begin{figure}[tb]
    \centering
    \includegraphics[scale=0.6]{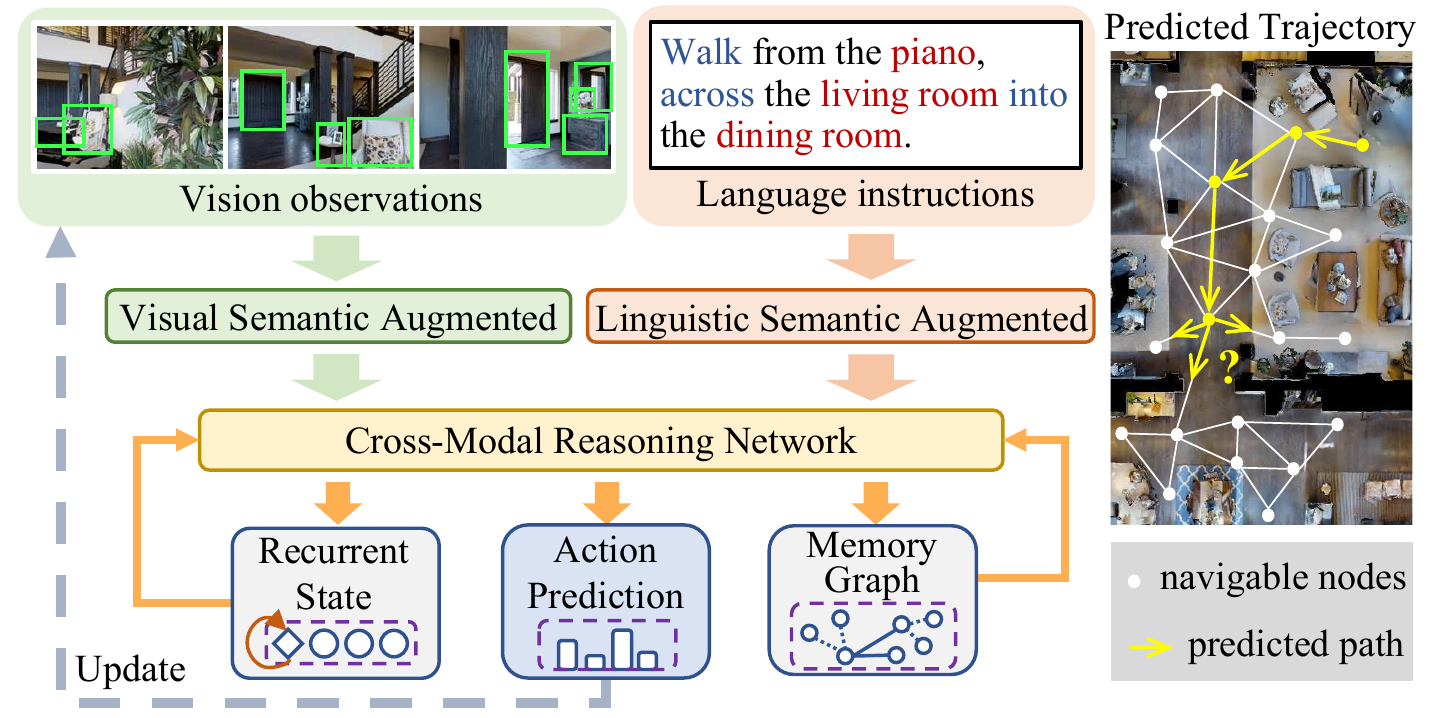}
    \caption{Illustration of the proposed DSRG, which first augments semantics of visual and linguistic inputs respectively, and employs the cross-modal reasoning network along with the recurrent state fusion module and the memory graph to predict the action.}
    \label{fig_intro}
\end{figure}

Due to the huge success of the transformer~\cite{vaswani2017attention}, transformer-based cross-modal fusion methods~\cite{hao2020towards,hong2021vln,zhu2020actbert,chen2022learning} have undergone substantial development and show promise. However, these methods still have some limitations. First, existing cross-modal networks don't leverage the guiding semantic information hidden in different inputs sufficiently. Intuitively, the guiding semantic features are both hidden in visual inputs (e.g., the object features in images) and linguistic inputs (e.g., the direction and landmark tokens). However, some recent semantic-related approaches~\cite{qi2021road,zhang2021diagnosing,an2021neighbor,dang2023multiple} either focus on the semantic enhancement of only one modality or employ several independent soft attention modules to learn correlation in special embeddings, without considering the simultaneous explicit modeling of finer visual and linguistic semantics for precise perception.

Secondly, the history-dependent ability of current models to infer the action for each step is inadequate. The long-term decision-making process requires the agent to keep track of the exploration progress with respect to its corresponding sub-instruction. While some recent transformer-based methods~\cite{chen2021history,chen2022think,chen2022reinforced} structurally model the historical information of visited nodes as the graph map, they simply provide average exterior features of past observations, lacking the flow of reasoning presentation within the network, which could lead to information loss and reasoning discontinuity.

\begin{figure*}[t]
    \centering
    \includegraphics[scale=0.58]{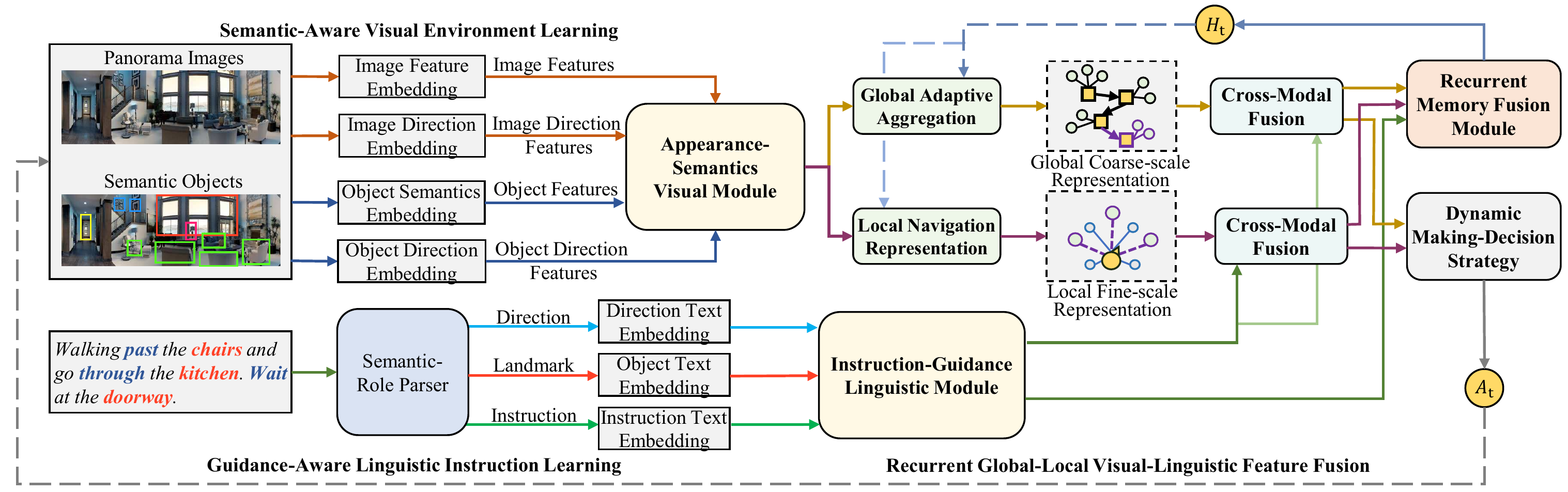}
    \caption{Overview of the proposed DSRG structure, which includes three components: (a) semantic-aware visual environment learning, (b) guidance-aware linguistic instruction learning, and (c) recurrent global-local visual-linguistic feature fusion.}
    \label{fig_overall}
\end{figure*}
To address the above problems, we propose DSRG, a novel dual semantic-aware recurrent global-adaptive network for VLN, where the guiding semantic features concealed in the visual and linguistic representations are better utilized to guide the navigation, and the memory augmentation is implemented based on the explicit outer flow (the structured graph of visited nodes) and the implicit inner flow (the specific recurrent memory token). To achieve these two goals, as shown in Fig.~\ref{fig_intro}, we first design a dual structure, including an instruction-guidance linguistic module (IGL) and an appearance-semantics visual module (ASV), to learn guiding semantic presentations of two types of inputs, respectively. Based on the dominant semantic prior, the agent is capable to capture the key components of visual and verbal inputs. Then, for the memory mechanism, we first adopt the global map used in~\cite{chen2022think} with a proposed global adaptive aggregation method (GAA) to fuse panoramic observations of visited nodes by adaptively learning the contribution of various views to candidate sites. Further, to improve the continuity of reasoning procedures within the network, we suggest a recurrent memory fusion module (RMF) to broadcast the inference states in a recurrent way. Experimental results on two datasets, R2R~\cite{anderson2018vision} and REVERIE~\cite{qi2020reverie}, have proved the effectiveness of our method, achieving a new state-of-the-art borderline on the VLN task. 

Overall, our contributions are summarised as follows: (1) We propose a dual semantic-augmented structure to boost visual and linguistic semantic representations, respectively. (2) We propose to use both explicit and implicit memory transfer channels for enhancing the model's ability to adaptively memorize and infer the status of navigation. (3) Extensive experiments on two datasets, R2R and REVERIE, demonstrate that the proposed DSRG outperforms other existing methods.

\section{Related Work}
\label{sec_related_work}
The vision-and-language navigation (VLN) task is first proposed by~\cite{anderson2018vision}, which serves as a new technique for relating natural language to vision and action in unstructured and unseen environments. 
To improve the generalization of the agent in unseen environments, some environment-augmented methods~\cite{fried2018speaker,tan2019learning,liu2021vision,li2022envedit,liang2022contrastive,wang2022res,wang2023pasts} are proposed for improving the diversity of scenes and instructions. Additionally, the task-specific auxiliary task methods~\cite{ma2019regretful,zhu2020vision,zhao2022target,he2023mlanet} are proposed to enhance the interpretability and navigation ability of the model. In general, most VLN methods encode visual and linguistic features via the large pre-trained networks, without considering the explicit usage of the semantic-level cues of the two types of inputs which are crucial for directing the agent.

\paragraph{Semantics in VLN.} Recently, some methods have demonstrated the advantages of semantic features for vision-based navigation~\cite{dang2022unbiased,dang2022search}. ORIST~\cite{qi2021road} and SOAT~\cite{moudgil2021soat} propose to concatenate object-level features with the scene-level features and learn them through the transformer encoder in a parallel way. BiasVLN~\cite{zhang2021diagnosing} observes that the low-level image features result in environmental bias. SEvol~\cite{chen2022reinforced} utilizes a graph-based method to construct relationships of objects. However, all of the above methods only boost semantics from the visual perspective and neglect guiding hints hidden in natural language instructions. OAAM~\cite{qi2020object} and NvEM~\cite{an2021neighbor} apply independent soft attention to text embeddings to learn the object and action representations of instructions, and the latter also considers the neighboring objects of candidates. ADAPT~\cite{lin2022adapt} suggests using the CLIP~\cite{radford2021learning} with an action prompt to improve the action-level modality alignment. Different from earlier methods that only focus on the semantics of a single modality or implicitly learn semantics based on soft attention with hidden states, we argue that the crucial guiding information exists in both vision and language and should be explicitly highlighted. Therefore, we propose a dual structure to enhance semantic features for vision and language presentations by injecting the extracted prior knowledge and then fusing them through the adaptively global-local cross-modal module.

\paragraph{Historical Memory in VLN.} It is crucial for agents to represent both the current and prior states while navigating. For LSTM-based methods~\cite{wang2020active,an2021neighbor,wang2021structured}, the long-term historical information is broadcast via the inherent hidden state of the network. With the enormous success made by the transformer~\cite{vaswani2017attention}, a growing number of models have lately achieved better performance based on the transformer structure. Some pre-trained models~\cite{hao2020towards,guhur2021airbert,liangVisualLanguageNavigationPretraining2022} are proposed to improve the model's capability of representation via the synthesized dataset. To provide the model with historical observations and improve inference capability, some methods~\cite{chen2021topological,chen2021history,chen2022think} focus on constructing the structured memory graph of visited nodes. However, these methods neglect to transmit the network's reasoning states at each step, which might lead to the interruption of the inference process. Inspired by RecBERT~\cite{hong2021vln}, we propose to improve the global memory graph adopted in the previous SOTA method~\cite{chen2022think} by a global adaptive aggregation method and then implement a recurrent fusion module to present the current reasoning states.

\begin{figure}[!htb]
    \centering
    \includegraphics[scale=0.7]{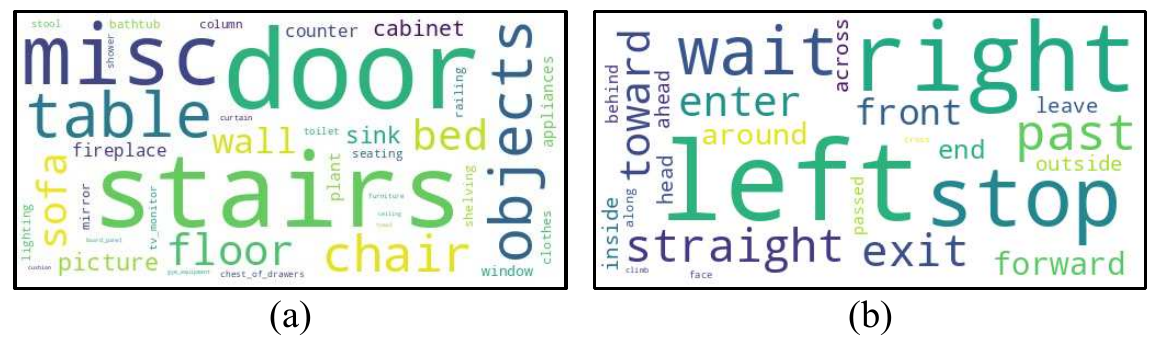}
    \caption{Word
    clouds of (a) landmark and (b) direction tokens.}
    \label{fig_cloudword}
\end{figure}
\label{sec_method}
\section{Methodology}
Based on the given instruction, the task of VLN is to predict sequence actions toward the target location automatically. Our goal is to take full advantage of the semantic features hidden in linguistic and visual representations to achieve more accurate navigation. As shown in Fig.~\ref{fig_overall}, the overall framework of our proposed method can be divided into three sub-modules: (a) guidance-aware linguistic instruction learning (Sec.~\ref{subsec_linguisitc_instruction_representation}), (b) semantic-aware visual environment learning (Sec.~\ref{subsec_visual_environment_representation}) and (c) recurrent global-local visual-linguistic feature fusion (Sec.~\ref{subsec_recurrent_global_local_visual_linguistic_feature_fusion}). Specifically, we first use a dual structure to enrich vision and language expressions by injecting guiding semantic features in their respective domains. Next, we construct a global navigation map via the adaptive fusion module to assist the local decision-making process. The inherent memory unit is updated based on the instruction, global map and local observations, and transferred to the next action step. Last, the possibility of action is calculated by the dynamic decision-making strategy.

\subsection{Guidance-Aware Linguistic Instruction Learning}
\label{subsec_linguisitc_instruction_representation}
The natural language contains a wealth of semantic information, serving as a vital link for social interaction. In this work, we first extract guiding semantic phrases from instructions and then encourage the model to focus on these dominant parts and enhance the instruction representations.

\paragraph{Semantic-Role Parser.} As shown in Fig.~\ref{fig_cloudword}, two kinds of semantic information are specified: direction-level and landmark-level phrases, where the former can guide the agent's forward direction and the latter can provide the referential landmark during navigation. While it is easy for humans to quickly capture these leading phrases, the network can only gradually learn this from massive data fitting. Therefore, we extract them based on the toolkit NLTK~\cite{nltk} to perform direction-landmark recognition based on its part-of-speech tags and entity dictionary. Following the category map provided by MP3D~\cite{chang2017matterport3d}, all extracted entities are normalized into 43 categories.
\begin{figure}[!tb]
    \centering
    \includegraphics[scale=0.62]{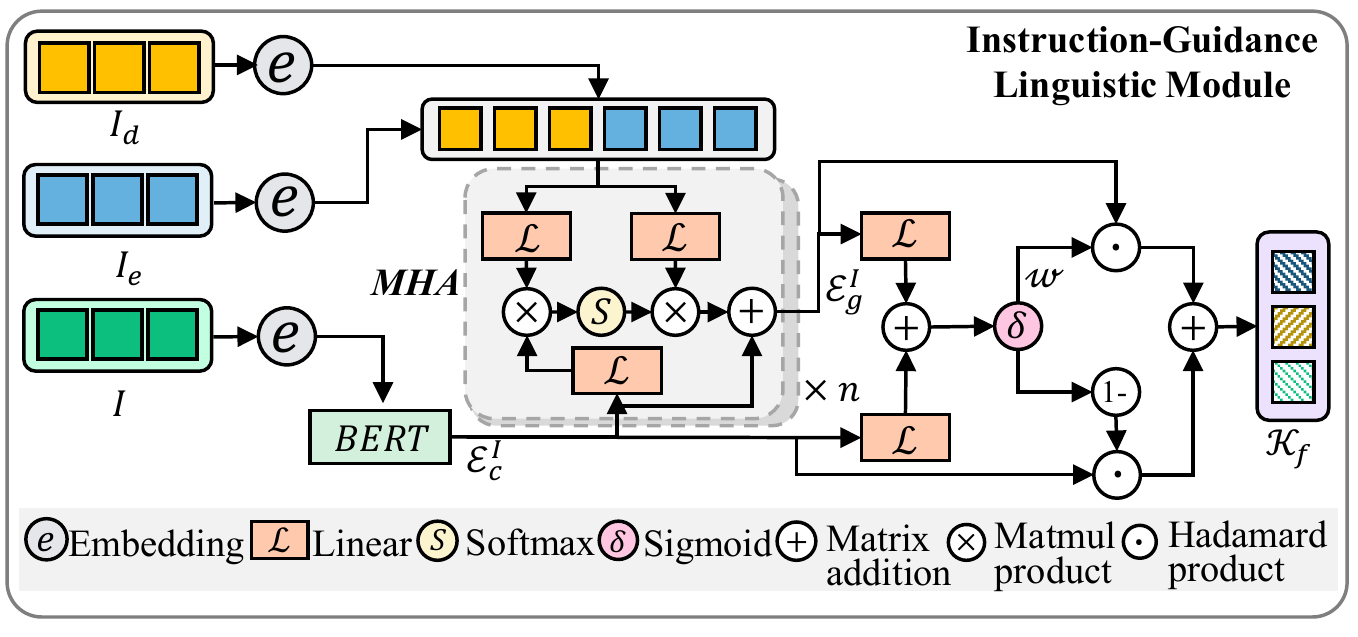}
    \caption{Illustration of the instruction-guidance linguistic module.}
    \label{fig_IGL}
\end{figure}
\paragraph{Instruction-Guidance Linguistic Module (IGL).} As shown in Fig.~\ref{fig_IGL}, for the whole instruction $I=\{w_1,...,w_L\}$, direction-level phrases $I_d=\{w^d_1,...,w^d_{L_d}\}$, and landmark-level phrases $I_e=\{w^e_1,...,w^e_{L_e}\}$, each token is firstly respectively embedded into a vector of 768-dimension. The position embedding function is employed to inject the ordering relation. Then the tokens in $I$ are encoded by a BERT~\cite{devlin2018bert} model to get context linguistic features denoted as $\mathcal{E}^{I}_{c}$. To highlight the guided representations of the key areas in the instruction as $\mathcal{E}^{I}_{g}$, we propose to use the multi-head attention (MHA) mechanism~\cite{vaswani2017attention} to update the context features based on the correlation between context tokens and the concatenate of direction and landmark tokens. Then, to balance the proportion of context content $\mathcal{E}_{c}^{I}$ and guided semantic features $\mathcal{E}_{g}^{I}$, we use a gate-like structure with a sigmoid function $\delta(\cdot)$ to dynamically obtain the weight value $\omega \in \mathbb{R}^{L\times 1}$ and then weighted sum the above two kinds of features into $\mathcal{K}_f \in \mathbb{R}^{L\times d_m}$. The above process is formulated as Eq.~\eqref{eq_IGL_1} --~\eqref{eq_IGL_2}:
\begin{align}
    \label{eq_IGL_1}
    (\mathcal{I},\mathcal{I}_d,\mathcal{I}_e) &= \text{Embedding}(I,I_d,I_e) \\
    \mathcal{E}_{c}^{I}&=\text{BERT}(\mathcal{I}) \\
    \mathcal{E}_{g}^{I} &= \text{MHA}(\mathcal{E}^{I}_{c}, [\mathcal{I}_{d},\mathcal{I}_{e}]) \\
    \omega &= \delta(\mathcal{E}^{I}_{g}W_g+\mathcal{E}_{c}^{I}W_c+b_I) \\
    \mathcal{K}_{f} &= \omega \odot \mathcal{E}^{I}_{g} + (1-\omega)\odot \mathcal{E}^{I}_{c} \label{eq_IGL_2}
\end{align}

where $[.,.]$ denotes concatenation operation, $W_g \in \mathbb{R}^{d_m \times 1}$ and $W_c \in \mathbb{R}^{d_m \times 1}$ present the learnable parameters.

\subsection{Semantic-Aware Visual Environment Learning}
\label{subsec_visual_environment_representation}
In VLN, another important modality of input is visual observation as the agent can only explore the unstructured environments based on images. Therefore, we further devise a semantic-aware visual environment learning approach to form a dual semantic-aware structure with the enhanced language features discussed in Sec.~\ref{subsec_linguisitc_instruction_representation}. By considering the fine-grained semantic object features within each sub-image of the panorama and fusing them with the image features, the environmental presentations obtain semantic enhancement. 

\paragraph{Visual Image Feature.} The connected navigation graphs are specified by the Matterport3D simulator~\cite{chang2017matterport3d}, where the navigable nodes are given discretely. Formally, each panorama is split into 36 images $V=\{v_i\}_{i=1}^{36}$. The corresponding heading $\theta$ and elevation $\gamma$ are denoted as $V_r=\{r_i\}_{i=1}^{36}$, where $r_i=(\sin \theta_i,\cos \theta_i,\sin \gamma_i,\cos \gamma_i) \in \mathbb{R}^{4}$. The ViT~\cite{dosovitskiy2020image} model is employed to extract image features. Aggregated by the angle features $V_{r} \in \mathbb{R}^{36\times 4}$, token types $V_{t} \in \mathbb{R}^{36\times 1}$ (initialized as ones), and navigable types $V_{n} \in \mathbb{R}^{36\times 1}$ (one for the navigable and zero for the non-navigable), the visual image features $\mathcal{E}^{V}_{f} \in \mathbb{R}^{36 \times d_m}$ are formulated as Eq.~\eqref{eq_VIF_1} --~\eqref{eq_VIF_2}:
\begin{align}
    \label{eq_VIF_1}
    [\mathcal{E}^{V}_{r},\mathcal{E}^{V}_{t},\mathcal{E}^{V}_{n}]&=[V_r,V_t,V_n]W_z+b_z \\
    \mathcal{E}^{V}_{v}&=\text{ViT}(V)W_v \\
    \mathcal{E}^{V}_{f} &= \text{LN}(\mathcal{E}^{V}_{v}+\mathcal{E}^{V}_{r}+\mathcal{E}^{V}_{t}+\mathcal{E}^{V}_{n})
    \label{eq_VIF_2}
\end{align}
where $W_z \in \mathbb{R}^{6 \times d_m},b_z \in \mathbb{R}^{d_m}, W_v \in \mathbb{R}^{d_v \times d_m}$.

\paragraph{Semantic Object Feature.} The previous methods~\cite{moudgil2021soat,qi2021road} simply concatenate objects and images in parallel without considering respective object features in 36 images of $V$, which may lead to the problem of feature dislocation. In contrast, we propose to use the fine-grained object features of 36 images for each viewpoint, which can be denoted as $O=\{[o_{i1},...,o_{iM}]\}_{i=1}^{36}$, where $M$ is the maximum number of objects in each image. Let $O_g \in \mathbb{R}^{36 \times M \times 3},O_r \in \mathbb{R}^{36 \times M \times 4},O_l \in \mathbb{R}^{36 \times M \times 1},O_n \in \mathbb{R}^{36 \times M \times 1}$ denote the geometric features, angle features, label features, and navigable types, respectively. To better leverage the semantic information, we filter out some less informative items (e.g., walls and ceilings) and take the first $M$ items with the largest area. The semantic object features $\mathcal{E}^{O}_f \in \mathbb{R}^{36 \times M \times d_o}$ are computed as follows:
\begin{align}
    [\mathcal{E}^{O}_{g},\mathcal{E}^{O}_{r},\mathcal{E}^{O}_{l},\mathcal{E}^{O}_{n}]&=[O_g,O_r,O_l,O_n]W_x+b_x \\
    \mathcal{E}^{O}_{f} &= \text{LN}(\mathcal{E}^{O}_{g}+\mathcal{E}^{O}_{r}+\mathcal{E}^{O}_{l}+\mathcal{E}^{O}_{n})
\end{align}
where $W_x \in \mathbb{R}^{9 \times d_o}$ and $b_x \in \mathbb{R}^{d_o}$ are learnable parameters.
\begin{figure}[tb]
    \centering
    \includegraphics[scale=0.67]{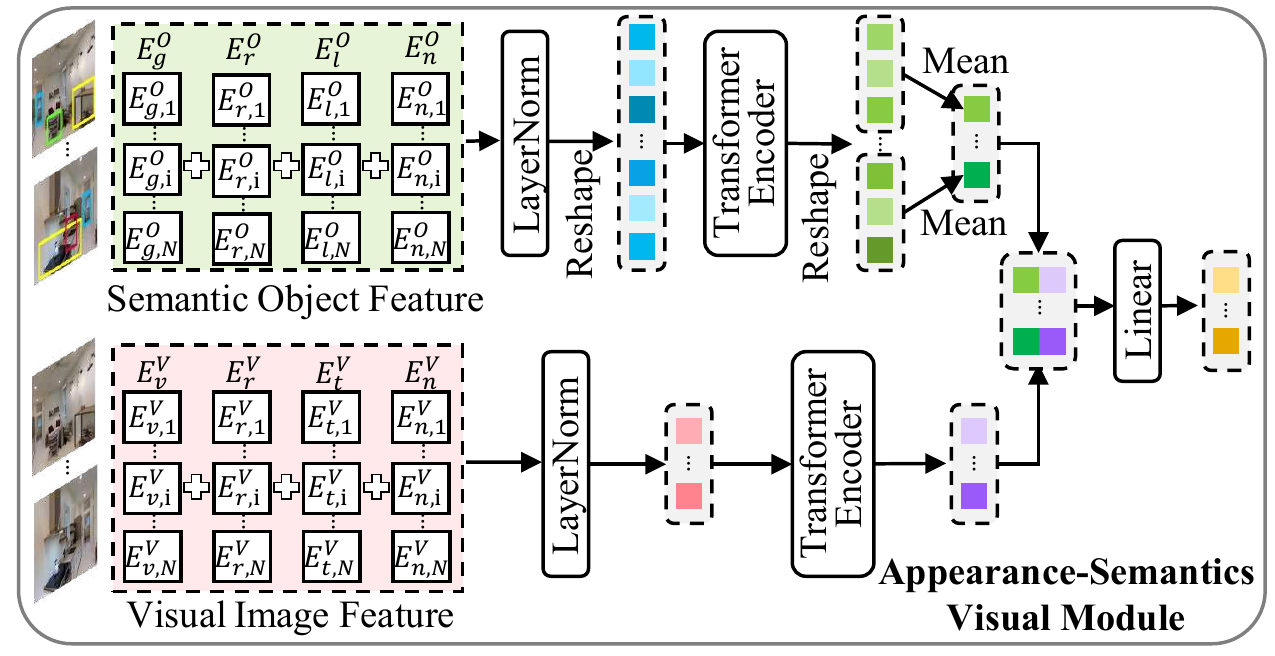}
    \caption{Illustration of the appearance-semantics visual module.}
    \label{fig_ASV}
\end{figure}
\paragraph{Appearance-Semantics Visual Module (ASV).} As there is a large overlap space between 36 images, the objects have natural internal relationships with each other as well. Therefore, as shown in Fig.~\ref{fig_ASV}, we first reshape object features from different images as $\widetilde{\mathcal{E}}^{O}_{f} \in \mathbb{R}^{(36\times M)\times d_o}$ and use a transformer block to promote intermediate semantic-level representations. Similarly, another transformer is used to learn the relationships between images, given by
\begin{align}
\widetilde{\mathcal{S}}^{O}_{f}=\text{Trans}_o(\widetilde{\mathcal{E}}^{O}_{f}), \, 
    \mathcal{S}^{V}_{f}=\text{Trans}_v(\mathcal{E}^{V}_{f})
\end{align}
where $\text{Trans}_o$ and $\text{Trans}_v$ mean the standard transformer block consisting of multi-head self-attention, residual connection, layernorm (LN) and feed-forward network (FFN). After obtaining the cross-image object feature $\widetilde{\mathcal{S}}^{O}_{l}$, we reshape it along the image dimension and then calculate the average representation of objects in each image. Finally, we concatenate the appearance-level features $\mathcal{S}^{V}_{f}$ and semantics-level features $\mathcal{S}^{O}_{l}$ together to obtain the semantic-aware visual representations as:
\begin{align}
    {\mathcal{S}}^{O}_{f}&=\frac{1}{M}\sum_{i=1}^{M}\widetilde{\mathcal{S}}^{O}_{f,i} \\
    \mathcal{S}_{f}&=\text{LN}([\mathcal{S}_{f}^{V},\mathcal{S}_{l}^{O}]W_f+b_f)
\end{align}
where $W_f \in \mathbb{R}^{(d_m+d_o) \times d_m},b_f \in \mathbb{R}^{d_m}$ are learnable parameters. In this work we employ $d_o=128$.

\subsection{Recurrent Global-Local Visual-Linguistic Feature Fusion}
\label{subsec_recurrent_global_local_visual_linguistic_feature_fusion}
The previous method~\cite{chen2022think} has shown the effectiveness of using a global memory graph based on average visual observations to promote inference capability. We argue that this is necessary but insufficient since it ignores the different contributions of images in panorama, and the reasoning states within the network have not got fully transferred through the navigation steps. Therefore, in this section, we devise a global adaptive aggregation method and a recurrent memory fusion module for memory augmentation.

\paragraph{Local Navigation Representation.} To provide fine-scale observation of the current node, the representations in the local branch are first comprised of enhanced semantic-aware visual features $\mathcal{S}_f$ described in Sec.~\ref{subsec_visual_environment_representation}, concatenated by a zero-initialized $\texttt{[CLS]}$ token (which also serves as the stop token) and the recurrent memory token $\texttt{[MEM]}$. The structure of the local navigation representation can be formulated as $\mathcal{U}_{L}=(\texttt{[CLS]}, \mathcal{S}_{f}, \texttt{[MEM]})$, where the calculation of \texttt{[MEM]} will be explained in the following. The embedding of the local position features is further added to $\mathcal{U}_{l}$ as $\widetilde{\mathcal{U}}_{l}$ to present the relative position information.

\paragraph{Global Adaptive Aggregation (GAA).} In DUET~\cite{chen2022think}, the global navigation map is constructed by storing visited nodes, navigable nodes, and the current node with their appearance features at each step, which has shown superior performance. Specifically, the visual representation of each node is represented by the average function $\frac{1}{N}\sum_{i=1}^{N}\mathcal{E}^{V}_{f,i}$, where $N$ denotes the length of tokens in each panoramic observation. However, this operation flattens the different contributions of different images. Intuitively, the images that do not contain specific landmarks or are far aware from candidate points have smaller contributions. It is necessary to pay more attention to more instruction-relevant parts to guide navigation. Therefore, we propose the GAA module to make the network adaptively learn to encode relationships between images. Supposed $\mathcal{S}_f=\{s_i\}_{i=1}^{N} \in \mathbb{R}^{N\times d_m}$, we introduce an attention matrix $W_f \in \mathbb{R}^{d_m \times 1}$ with the softmax function to adaptively re-weight image features and sum them up, achieving the aggregated features $\mathcal{S}_{a}\in \mathbb{R}^{1\times d_m}$:
\begin{align}
    \mathcal{R}=[r_1,...,r_N]&=\gamma(\mathcal{S}_fW_f+b_f) \\
    \widetilde{r}_i&=\frac{e^{r_i}}{\sum_{j\in N}e^{r_j}} \\
    \mathcal{S}_a&=\sum_{i\in N}\widetilde{r}_i * s_i 
\end{align}
where $\gamma(\cdot)$ is the tanh activation function. Let $\mathcal{S}_{g}$ denote the set of aggregated node features in the global map, similar to the local branch, the structure of the global map representation is formulated as $\mathcal{U}_{g}=(\texttt{[CLS]},\mathcal{S}_{g},\texttt{[MEM]})$. The embedding of step position is also added to $\mathcal{U}_g$ as $\widetilde{\mathcal{U}}_g$.

\paragraph{Cross-Modal Feature Fusion.} After obtaining the semantic-augmented features for both vision and language, it is essential to fuse these features together to learn the correlation between these two kinds of inputs. LXMERT~\cite{tan2019lxmert} is adopted as the cross-modal encoder. Following~\cite{hong2021vln}, the language tokens are only assigned as keys and values to update the visual tokens during the fine-tuning.
Concretely, two independent cross-modal fusion modules are employed to fuse the linguistic features $\mathcal{K}_f$ with the global and local visual features $\{\widetilde{\mathcal{U}}_g,\widetilde{\mathcal{U}}_l\}$, achieving $\mathcal{F}_g$ and $\mathcal{F}_l$, respectively.

\paragraph{Recurrent Memory Fusion Module (RMF).} 
\label{para_rmf}
Considering the construction of the global map is an explicit memory representation, we propose to further build an RMF module to transmit the network's intermediate reasoning states. Intuitively, in the sequential reasoning task, we can always remember the basis of our own judgment at the last moment, which can make the subsequent reasoning easier and more accurate. Therefore, for the $t$-th step, we extract the $\texttt{[CLS]}$ tokens from $\mathcal{F}_g$, $\mathcal{F}_l$ and $\mathcal{K}_f$ as $\mathcal{C}_{g},\mathcal{C}_{l}$ and $\mathcal{C}_{k}$, respectively. This is because $\texttt{[CLS]}$ tokens can be regarded as the highly fused presentations of the corresponding modalities at the current step. Then we project them to the memory representation domains via the linear transformations after concatenating:
\begin{align}
    \mathcal{H}_{r}&=\text{LN}([\mathcal{C}_{g},\mathcal{C}_{l},\mathcal{C}_{k}]W_c+b_c)
\end{align}
where $W_c \in \mathbb{R}^{3d_m\times d_m}$ and $b \in \mathbb{R}^{d_m}$. To avoid interference in the prediction of the stop signal, a separate token $\texttt{[MEM]}$ is defined to specifically store the hidden reasoning representation in each step. The obtained inherent memory unit $\mathcal{H}_{r} \in \mathbb{R}^{1\times d_m}$ is assigned by the recurrent token $\texttt{[MEM]}$ of the global and local sequence at the next step. 

\paragraph{Dynamic Decision-Making Strategy.} Finally, we employ the dynamic decision-making strategy proposed by~\cite{chen2022think} to predict the action.  
Concretely, the one-dimensional scalar weight $\sigma$ of global and local branches are computed by the FFN network on the concatenation of $\mathcal{C}_g$ and $\mathcal{C}_l$, respectively. After using another two FFN networks to project global-local fused features into the score domains, the local action scores $\mathcal{G}_{l}$ are converted into the global action space $\hat{\mathcal{G}}_{l}$. Then the final probability of the action prediction $\mathcal{G}$ is obtained via the weighted sum of the two branches. Only candidate nodes will be considered based on the mask function. The formulas are as Eq.~\eqref{eq_decisionmaking_1} --~\eqref{eq_decisionmaking_2}:
\begin{align}
\label{eq_decisionmaking_1}
    \sigma&=\delta(\text{FFN}([\mathcal{C}_g,\mathcal{C}_l]))\\
    \mathcal{G}_g&=\text{FFN}(\mathcal{F}_g),\,\mathcal{G}_l=\text{FFN}(\mathcal{F}_l) \\
    \mathcal{G}&=\sigma \mathcal{G}_g + (1-\sigma) \hat{\mathcal{G}}_l \label{eq_decisionmaking_2}
\end{align}
The cross-entropy loss is used as the optimization objective:
\begin{equation}
    \mathcal{L}=\sum_{t=1}^T-\log p(a_t^*|\mathcal{I},\mathcal{V}_t,a_{1:t-1})
\end{equation}

\begin{table*}[!htp]
\centering
\resizebox{\linewidth}{!}{
\begin{tabular}{l|rrrr|rrrr|rrrr}
\toprule
\multirow{2}{*}{Methods} & \multicolumn{4}{c|}{Validation Seen} & \multicolumn{4}{c|}{Validation Unseen} & \multicolumn{4}{c}{Test Unseen} \\ 
  & NE$\downarrow$ & OSR$\uparrow$ & SR$\uparrow$ & SPL$\uparrow$ & NE$\downarrow$ & OSR$\uparrow$ & SR$\uparrow$ & SPL$\uparrow$ & NE$\downarrow$ & OSR$\uparrow$ & SR$\uparrow$ & SPL$\uparrow$ \\ \midrule
EnvDrop~\cite{tan2019learning} & 3.99 & - & 62 & 59 & 5.22 & - & 52 & 48 & 5.23 & 59 & 51 & 47 \\
AuxRN~\cite{zhu2020vision} & 3.33 & 78 & 70 & 67 & 5.28 & 62 & 55 & 50 & 5.15 & 62 & 55 & 51 \\
PREVALENT~\cite{hao2020towards} & 3.67 & - & 60 & 65 & 4.73 & - & 57 & 53 & 4.75 & 61 & 54 & 51 \\
RecBERT~\cite{hong2021vln} & 2.90 & 79 & 72 & 68 & 3.93 & 69 & 63 & 57 & 4.09 & 70 & 63 & 57 \\
NvEM~\cite{an2021neighbor} & 3.44 & - & 69 & 65 & 4.27 & - & 60 & 55 & 4.37 & 66 & 58 & 54 \\
HOP~\cite{qiao2022hop} & 2.72 & - & 75 & 70 & 3.80 & - & 64 & 57 & 3.83 & - & 64 & 59 \\
EnvMix~\cite{liu2021vision}& 2.48 & - & 75 & 72 & 3.89 & - & 64 & 58 & 3.87 & 72 & 65 & 59 \\
HAMT~\cite{chen2021history} & 2.51 & 82 & 76 & 72 & \textbf{2.29} & 73 & 66 & 61 & 3.93 & 72 & 65 & 60 \\
TD-STP~\cite{zhao2022target} & 2.34 & 83 & 77 & 73 & 3.22 & 76 & 70 & \textbf{63} & 3.73 & 72 & 67 & 61 \\
DUET~\cite{chen2022think} & 2.28 & 86 & 79 & 73 & 3.31 & 81 & 72 & 60 & 3.65 & 76 & 69 & 59 \\ 
\hline
\textbf{DSRG (Ours)} & \textbf{2.23} & \textbf{88} & \textbf{81} & \textbf{76} & 3.00 & \textbf{81} & \textbf{73} & 62 & \textbf{3.33} & \textbf{78} & \textbf{72} & \textbf{61} \\ \bottomrule
\end{tabular}}
\caption{Comparison with the state-of-the-art methods on the R2R dataset.}
\label{tab_r2r}
\end{table*}
\begin{table*}[!htb]
\centering
\setlength\tabcolsep{1pt}
\renewcommand{\arraystretch}{1.1}
\resizebox{\linewidth}{!}{
\begin{tabular}{l|ccccc|ccccc|ccccc}
\toprule
\multirow{3}{*}{Methods} & \multicolumn{5}{c|}{Val Seen} & \multicolumn{5}{c|}{Val Unseen} & \multicolumn{5}{c}{Test Unseen} \\
 & \multicolumn{3}{c}{Navigation} & \multicolumn{2}{c|}{Grounding} & \multicolumn{3}{c}{Navigation} & \multicolumn{2}{c|}{Grounding} & \multicolumn{3}{c}{Navigation} & \multicolumn{2}{c}{Grounding} \\
 & OSR$\uparrow$ & SR$\uparrow$ & SPL$\uparrow$ & RGS$\uparrow$ & RGSPL$\uparrow$ & OSR$\uparrow$ & SR$\uparrow$ & SPL$\uparrow$ & RGS$\uparrow$ & RGSPL$\uparrow$ & OSR$\uparrow$ & SR$\uparrow$ & SPL$\uparrow$ & RGS$\uparrow$ & RGSPL$\uparrow$ \\ \midrule
RCM~\cite{wang2019reinforced} & 29.44 & 23.33 & 21.82 & 16.23 & 15.36 & 14.23 & 9.29 & 6.97 & 4.89 & 3.89 & 11.68 & 7.84 & 6.67 & 3.67 & 3.14 \\
MATTN~\cite{qi2020reverie} & 55.17 & 50.53 & 45.50 & 31.97 & 29.66 & 28.20 & 14.40 & 7.19 & 7.84 & 4.67 & 30.63 & 19.88 & 11.61 & 11.28 & 6.08 \\
SIA~\cite{Lin_2021_CVPR} & 65.85 & 61.91 & 57.08 & 45.96 & 42.65 & 44.67 & 31.53 & 16.28 & 22.41 & 11.56 & 44.56 & 30.80 & 14.85 & 19.02 & 9.20 \\
RecBERT~\cite{hong2021vln} & 53.90 & 41.79 & 47.96 & 38.23 & 35.61 & 35.02 & 30.67 & 24.90 & 18.77 & 15.27 & 32.91 & 29.61 & 23.99 & 16.50 & 13.51 \\
HOP~\cite{qiao2022hop} & 54.88 & 53.76 & 47.19 & 38.65 & 33.85 & 36.24 & 31.78 & 26.11 & 18.85 & 15.73 & 33.06 & 24.34 & 16.38 & 17.69 & 14.34 \\
HAMT~\cite{chen2021history} & 47.65 & 43.29 & 40.19 & 27.20 & 25.18 & 36.84 & 32.95 & 30.20 & 18.92 & 17.28 & 33.41 & 30.40 & 26.67 & 14.88 & 13.08 \\
DUET~\cite{chen2022think} & 73.86 & 71.75 & 63.94 & 57.41 & 51.14 & 51.07 & 46.98 & 33.73 & 32.15 & 23.03 & 56.91 & 52.51 & 36.06 & 31.88 & 22.06 \\ \hline
\multicolumn{1}{l|}{\textbf{DSRG (Ours)}} & \textbf{77.72} & \textbf{75.69} & \textbf{68.09} & \textbf{61.07} & \textbf{54.72} & \textbf{53.25} & \textbf{47.83} & \textbf{34.02} & \textbf{32.69} & \textbf{23.37} & \textbf{58.26} & \textbf{54.04} & \textbf{37.09} & \textbf{32.49} & \textbf{22.18} \\ \bottomrule
\end{tabular}}
\caption{Comparison with the state-of-the-art methods on the REVERIE dataset.}
\label{tab_reverie}
\end{table*}

\begin{table}[!htp]
\centering
\setlength{\tabcolsep}{1.8mm}{
\begin{tabular}{l|cc|rrrr}
\toprule
Id & ASV & IGL & SR$\uparrow$ & SPL$\uparrow$ & NE$\downarrow$ & OSR$\uparrow$ \\ \bottomrule
1 & \ding{55} & \ding{55} & 69.18 & 60.28 & 3.36 & 77.31 \\
2 & \ding{51} & \ding{55}  & 71.43 & 61.07 & 3.37 & 78.37 \\
3 & \ding{55} & \ding{51} & 71.26 & 61.20 & 3.26 & 78.93 \\
\textbf{4} & \textbf{\ding{51}} & \textbf{\ding{51}} & \textbf{72.50} & \textbf{61.56} & \textbf{3.00} & \textbf{80.97} \\ \bottomrule
\end{tabular}}
\caption{Ablation study for the dual semantic-aware modules.}
\label{tab_ablation_readnet}
\end{table}
\section{Experiments}
\label{sec_experiments}
\subsection{Dataset and Implementation Details}
\paragraph{Dataset.} To validate our proposed method, we conduct extensive experiments on the R2R~\cite{anderson2018vision} and REVERIE datasets~\cite{qi2020reverie}. Based on 90 different buildings, R2R includes 21,576 fine-grained step-by-step instructions to guide the agent, and REVERIE includes 21,702 shorter annotated instructions for remoted referring expressions. The fine-grained object features are provided by REVERIE, where each panoramic sub-image contains an average of 10 objects.

\paragraph{Evaluation Metrics.} For R2R, four standard metrics are for evaluation: the navigation error (NE): the distance between the ground truth and the agent's stop position; the success rate (SR): the ratio of paths that stop within 3m from the target points; the oracle success rate (OSR): SR with the oracle stop policy; and the success rate weighted by the path length (SPL): SR penalized by the path length. For REVERIE, another two metrics are added: remote grounding success rate (RGS): the ratio of objects grounded correctly, and the RGS weighted by the path length (RGSPL). 

\paragraph{Implementation Details.} In the pre-training stage, we train our DSRG with batch size 24 for 400k iterations using 1 NVIDIA RTX 3090 GPU. Since the supplement of directional and semantic texts will cause the leakage for the masked language modeling (MLM)~\cite{devlin2018bert}, and the recurrent state cannot be directly used for the single-step action prediction (SAP)~\cite{chen2021history}, we did not enable IGL and RMF in the pre-training phase, but added them in the fine-tuning stage. During fine-tuning, the batch size and the learning rate are 4 and $5\times 10^{-6}$, respectively. We use the ViT-B/16~\cite{dosovitskiy2020image} pre-trained model to extract image features. The numbers of transformer layers for instructions, visual and semantic features, and local-global cross-modal attention modules are 9, 2 and 4, respectively.

\subsection{Comparison with State-of-the-Arts}
The quantitative performance results in comparison with state-of-the-art methods on R2R and REVERIE datasets are listed in Table~\ref{tab_r2r} and Table~\ref{tab_reverie}, respectively. The proposed DSRG achieves the leading performance on both two datasets. Specifically, our model outperforms the current SOTA~\cite{chen2022think} significantly, with SPL being improved by 2\% and 2\% on the R2R unseen validation and test splits, respectively. The SR on REVERIE also obtains large improvements by about 4\% and 3.5\% on the seen and unseen splits, respectively. These results demonstrate that our model is beneficial for enhancing the performance of the agent in VLN, with a more fine-grained perception of the environmental and linguistic semantics and better long-term exploration understanding.
\begin{figure*}[tb]
    \centering
    \includegraphics[scale=0.62]{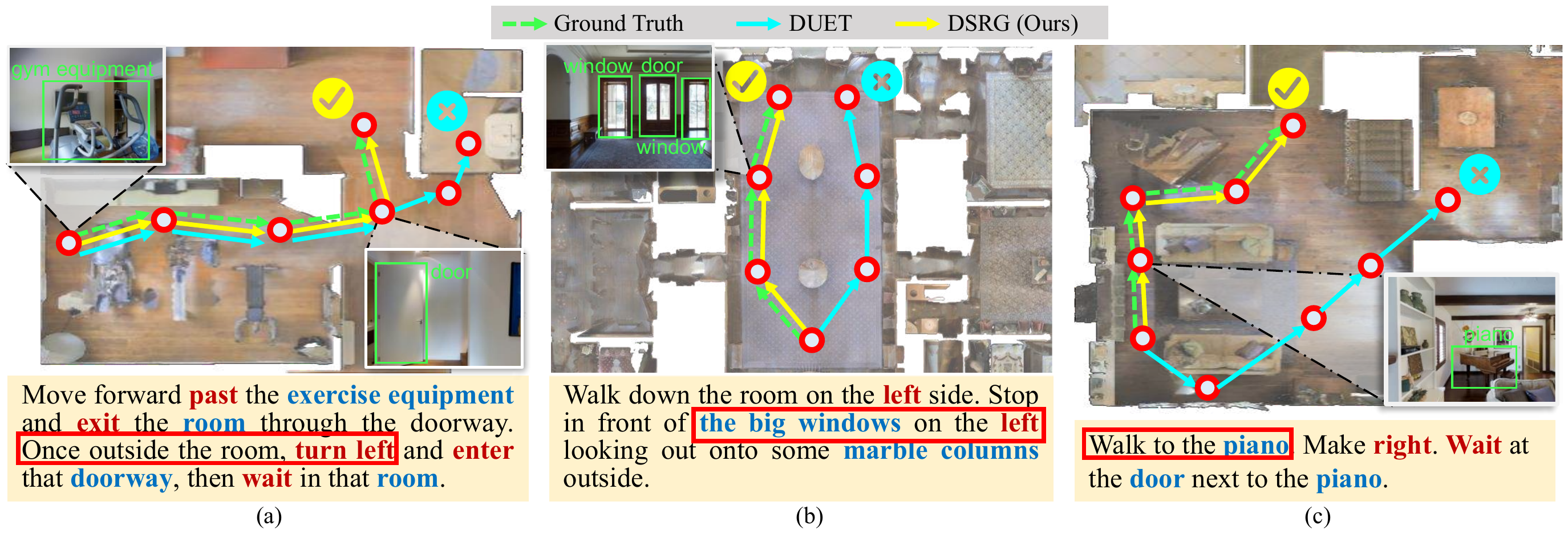}
    \caption{Visualization of navigation paths. The picture beside contains key guidance items which are marked in green. Below are the given language instructions, in which we use red for directions and blue for landmarks. Important phrases are highlighted by red boxes.}
    \label{fig_vis_path}
\end{figure*}

\subsection{Ablation Study}
We perform ablation studies to evaluate the proposed components of our method on the R2R unseen dataset.

\paragraph{Analysis on the Dual Semantic-Aware Modules.} Table~\ref{tab_ablation_readnet} shows several ablation studies to verify the effectiveness of our proposed dual semantic-aware modules consisting of ASV and IGL. Specifically, ASV and IGL are proposed to improve the semantic representations hidden in vision and language, respectively. We can observe that ASV and IGL are beneficial to effectively enhance the performance on all metrics alone, and this improvement is much more pronounced when these two modules are combined, with gains of 3.3\% on SR and 1.3\% on SPL. This demonstrates that with the help of semantics augmentation on both modalities, the model is capable to understand environments and instructions better.

\paragraph{Analysis on the Augmented Memory Mechanism.} To enhance memory representations through navigation steps, we propose 
GAA and RMF modules to adaptively aggregate the explicit global features of sub-images for each viewpoint, and transmit implicit reasoning states for each step, respectively. As shown in Table~\ref{tab_aggregation}, for the graph memory (denoted as ``GM''), GAA enhances all metrics significantly, demonstrating that our GAA method enables the model to assign weights reasonably by learning the different contributions of images based on their locations and contents. Additionally, for the recurrent states (denoted as ``RS''), we explore the effects of different branches on the RMF. It shows that the reasoning clues are simultaneously dependent on global-local features and language instructions. Overall, the agent can perform and leverage the history-dependent inference capabilities more effectively with the proposed GAA and RMF modules.
\begin{table}[!tp]
\centering
\setlength{\tabcolsep}{1.5mm}{
\begin{tabular}{ccl|rrrr}
\toprule
\multicolumn{3}{c|}{Method} & SR$\uparrow$ & SPL$\uparrow$ & NE$\downarrow$ & OSR$\uparrow$ \\ \bottomrule
\multicolumn{1}{c|}{\multirow{2}{*}{GM}} & \multicolumn{2}{c|}{w/o GAA} & 71.39 & 61.24 & 3.22 & 79.95 \\
\multicolumn{1}{c|}{} & \multicolumn{2}{c|}{\textbf{w/ GAA}} & \textbf{72.50} & \textbf{61.56} & \textbf{3.00} & \textbf{80.97} \\ \hline \hline
\multicolumn{1}{c|}{\multirow{5}{*}{RS}} & \multicolumn{2}{c|}{w/o RMF} & 72.16 & 60.73 & 3.14 & 80.29 \\ \cline{2-7} 
\multicolumn{1}{c|}{} & \multicolumn{1}{c|}{\multirow{4}{*}{w/ RMF}} & w/o global & 72.35 & 61.04 & 3.10 & 80.50 \\
\multicolumn{1}{c|}{} & \multicolumn{1}{c|}{} & w/o local & 71.18 & 60.89 & 3.13 & 80.25 \\
\multicolumn{1}{c|}{} & \multicolumn{1}{c|}{} & w/o text & 70.63 & 60.34 & 3.25 & 78.67 \\
\multicolumn{1}{c|}{} & \multicolumn{1}{c|}{} & \textbf{Full} & \textbf{72.50} & \textbf{61.56} & \textbf{3.00} & \textbf{80.97} \\ \bottomrule
\end{tabular}}
\caption{Ablation study for the memory mechanism.}
\label{tab_aggregation}
\end{table}

\begin{figure}[tb]
    \centering
    \includegraphics[scale=0.43]{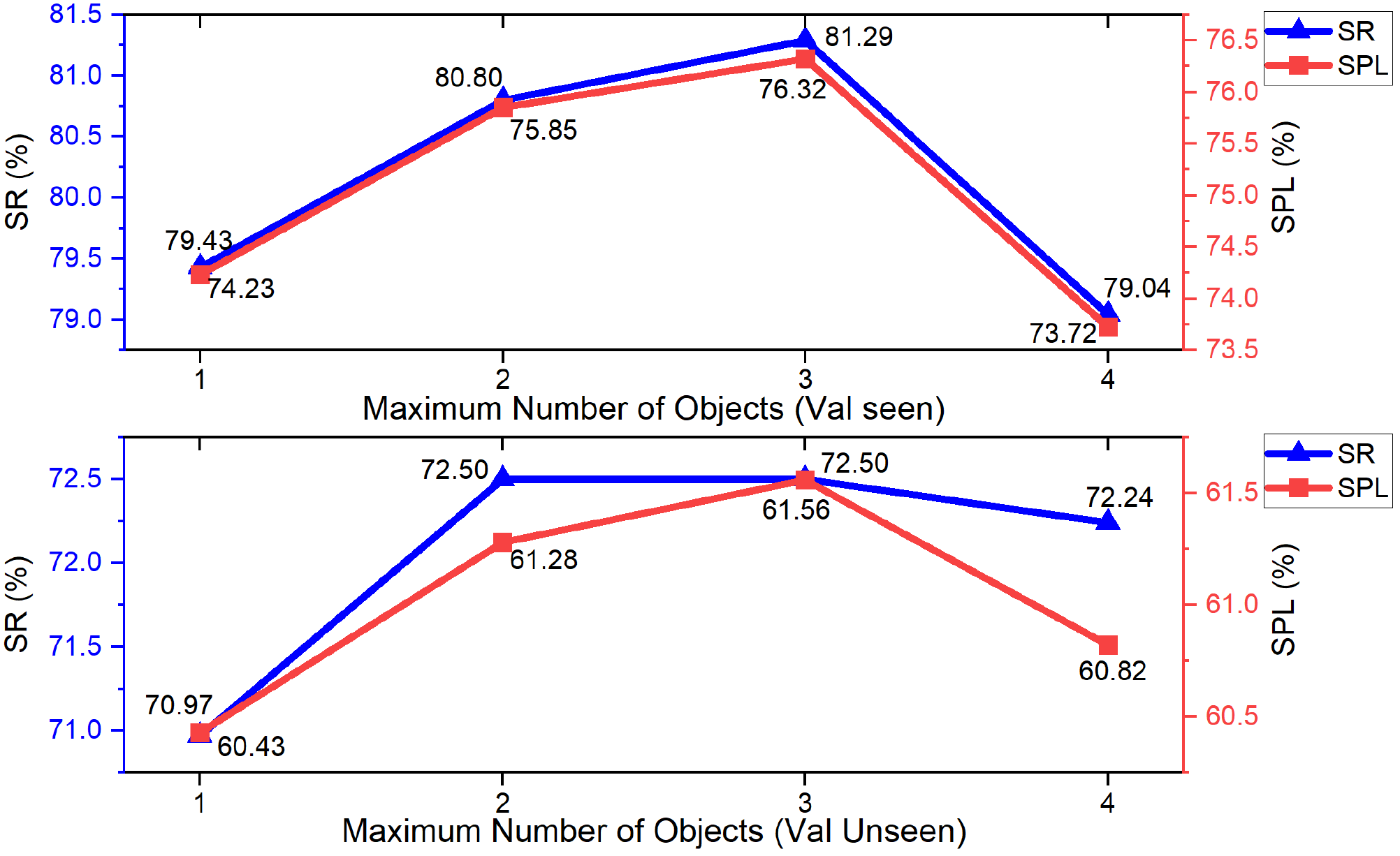}
    \caption{Ablation study for the maximum number of objects.}
    \label{fig_object_number}
\end{figure}

\paragraph{Analysis on the Number of Objects.} As described in Sec.~\ref{subsec_visual_environment_representation}, the cross-image relationships of objects are first learned by a transformer encoder block. Therefore, the maximum number of objects in each image will influence the lengths of tokens. Fig.~\ref{fig_object_number} shows that the model learning is optimal when the number of objects is 3 (totaling 108 objects for the current node). This demonstrates that too many objects will make the parallel lengths too long to pay attention to leading object features, while too few objects will result in a misinterpretation of visual semantics.

\subsection{Qualitative Results and Visualization}
Some visualization results are presented in this subsection to further analyze how our model contributes to the action decision during navigation. From Fig.~\ref{fig_vis_path} (a) we can see that our DSRG successfully seizes the moment to ``turn left", that is, ``once outside the room". However, DUET turns left late and enters the bathroom incorrectly. This indicates that our DSRG can better recognize the navigation progress and response for the explicit landmark. As for (b) and (c), our DSRG successfully chooses the right navigable nodes with respect to the ``big windows on the left" and the ``walk to the piano", which leads to correct navigation results, while DUET fails. This proves that a deeper understanding of the semantics both in instructions and observations is crucial for the VLN agent.

\section{Conclusion and Future Work}
In this paper, we present a dual semantic-aware recurrent global-adaptive network, namely DSRG, to improve the performance of agents in VLN. It can effectively recognize the guiding semantic information hidden in both linguistic instructions and visual observations with the help of the proposed ASV and IGL modules, respectively. For the memory mechanism, the GAA module is proposed to adaptively aggregate different sub-images in the panorama for global map construction. The RMF module is devised to supply implicit temporal hidden states by transferring reasoning cues from previous steps. Extensive experiments on two public datasets, R2R and REVERIE, have demonstrated that our method outperforms the state-of-the-art methods. With good expansibility and robustness, our approach is believed to have the potential to serve other VLN-like tasks as well, and we leave this exploration for future work.

\section*{Acknowledgments}
This paper is supported by the National Natural Science Foundation of China under Grants (62233013, 62073245, 62173248). Suzhou Key Industry Technological Innovation-Core Technology R\&D Program (SGC2021035); Special funds for Jiangsu Science and Technology Plan (BE2022119). Shanghai Municipal Science and Technology Major Project (2021SHZDZX0100) and the Fundamental Research Funds for the Central Universities. Shanghai Science and Technology Innovation Action Plan (22511104900).

\bibliographystyle{named}
\bibliography{ijcai23_abbr}

\end{document}